# A Predictive Services Architecture for Efficient Airspace Operations

Ítalo Romani de Oliveira, Samet Ayhan, Glaucia Balvedi, Michael Biglin, Pablo Costas
The Boeing Company

Euclides C. Pinto Neto
IEEE Member

Alexandre Leite, Felipe C. F. de Azevedo
MWF Services

***Abstract* — Predicting air traffic congestion states and flow management measures with certain degrees of confidence is a much desirable capability for airlines and Air Navigation Service Providers (ANSP), since this will result in better operational plans and efficiency. Estimates of future airport capacity and airspace density are crucial for a better managed airspace, both strategically and tactically, contributing to reduction in air traffic controller's workload and fuel consumption, ultimately leading to a more sustainable aviation. Such predictive capabilities have been subject of extensive literature and, although the prior approaches have been able to address these problems to some extent, the data management and query processing aspects remain more challenging than ever. With a vast and ever-increasing volume of air traffic data, generated at high rates, several analytics use-cases cannot be solved by ad-hoc approaches and need common pre-processing infrastructure. Also, linear prediction models fall short in most cases and more powerful techniques are in short order.**

**In this paper, we propose a data processing and predictive services architecture which ingests big, uncorrelated and noisy streaming data, and predicts future states of the airspace system. In the preprocessing step, the system continuously gathers the incoming raw data and, periodically, reduces it to a compact size and stores it in NoSQL databases, where the data becomes available for efficient query processing. In the prediction use cases, the system learns from historical traffic by collecting key features such as airport arrival and departure events, sector boundary crossings, weather parameters, and other air traffic data. These features are fed into various regression models, including linear, non-linear and ensemble models, and the best performing model is used for prediction. Evaluations of this infrastructure is performed in three prediction use-cases both in the US National Airspace System (NAS) and in a portion of the European airspace, with extensive sets of real operations data, and we verify that our system can predict efficiently and accurately future system states.**

## I. INTRODUCTION TO THE PROBLEMS TO BE SOLVED

In spite of having periods of retraction, the long-term growth trend of air traffic seems to be unstoppable. The US Federal Aviation and Administration (FAA) Office of Aviation Policy and Plans (APO) analysis of pre-pandemic years showed that the cost of delayed flights was consistently in the order of tenths of billions of dollars and, as of this writing, the global traffic levels are approaching 2019 figures. Most of the delay cost is due to inefficiencies in the major regional airspace systems around the globe, rooted in inherent capacity limitations. The inability to accurately predict future states of airspace and airports compounds the problem, resulting in added costs to the airlines and passengers [*1*]. Notwithstanding, forecasts of the aviation industry estimate that the demand for passenger-km will roughly double between now and 2040 [*2*,*3*], a challenge that becomes even more daunting with the industry commitment to become net zero in carbon emissions by 2050 [*4*]. Thus, besides having to recur to alternative sources of energy for aviation, efficiency improvements are of paramount importance, and the infrastructure capacity to handle flights has a prominent role in attaining efficiency.

Airport and airspace capacity management has been subject of studies since decades earlier, especially from the Economics Science point of view: [*5*] highlights the relationship of peak-time congestion with delays, reduced safety, and propose a slot pricing method to reduce delays; [*6*] further elaborates on congestion models and compares strategies of intertemporal adjustments and stochastic queuing; [*7*] highlights the need to adjust congestion pricing according to the degree of monopolization that an airline has in an airport; [*8*] introduces a comprehensive modelling framework for modelling congestion charges, based on diffusion theory; [*9*] focuses on arrival capacity and demonstrates that modest changes in flight patterns can reduce delays and congestion fees quite considerably. These studies assume that each airport has an intrinsic capacity, so what they set out to do is to balance the existing capacity among the several users. But can airport capacity be expanded? To a certain extent, yes, either by construction or by technology.

As airport construction is an expensive endeavor, Air Traffic Management (ATM) is in constant evolution [*10*]. New ATM concepts of operations include Trajectory-Based Operations (TBO), Collaborative Decision-Making (CDM), Dynamic Re-sectorization, Free-Flight sectors, and other innovations, all with the aim of improving air transportation efficiency. The realization of such concepts should enable higher degrees of automation and predictability, resulting in lower levels of human operator workload, and more streamlined operations. In this paper, we focus on predictive capabilities, which can be used for dynamically balancing demand to capacity, which is crucial for effciency.

## II. MEASURING AIR TRAFFIC CAPACITY

The capacity of the air transportation system is dictated by its most elementary physical resources, namely airspace sectors and airports. Although the technical and human constraints are highly significant, physical airspace can never be expanded, and airports compete for land with other societal needs, thus they are pretty much immutable in the short term and constitute the two major focus areas of capacity management.

### *A. Airport Capacity*

An airport's utilization depends on its capacity, and determination of airport capacity is a multidisciplinary science, which aims at providing estimates on how many inbound and outbound an airport can safely process at a given period of time. It is inherently inexact, since airport traffic is managed by humans, which manage human-piloted aircraft.

To begin with, the concept of airport capacity has several meanings, from which we will consider just a few:

- *Theoretical capacity*, which is a pair of inbound and outbound aircraft rates, resulting from the combination of infrastructure, current weather, and human traffic control team, under which the probability of safety incidents is acceptably low. To speak of *airport capacity* is to speak of a summarized quantity, because an airport can have multiple runway configurations, with different capacities, and the classification of the weather condition can be overly simplifying. There models with different degrees of realism for theoretical airport capacity, some more analytical [*11*,*12*,*13*], and some more relying on computational models for simulation [*14*,*15*].
- *Declared capacity*, which arises because of the uncertainties in knowing the theoretical capacity. This is the value communicated by the airport operator to other stakeholders of the air transportation system, which this entity commits to process and which constitutes its responsibility. Because of this responsibility aspect, it is subject to standards and regulations from aviation authorities [*16*,*17*,*18*,*19*].
- *Realized capacity*, which is simply the number of aircraft manifestly processed in a period of time, accounted after that period. Although this concept is easy to grasp, it can be misleading, because it is associated with a demand pressure [*20*]. If a well paved and equipped airport receives one aircraft per day or less, there is very little demand pressure and the realized capacity will not be representative of the theoretical capacity, hence devoid of meaning. Thus, realized capacity makes sense only when observing historical data, finding out the time periods with the largest realized rates of aircraft, and checking that in those periods there is concomitant occurrences of delay due to the impossibility of that airport processing more flights, signaling demand pressure. Furthermore, a statistically significant number of such occurrences has to be available in order to obtain reliable values of realized capacity; this is done in [*21*]. Factors such as the level of saturation at arrival processes, time of day, and meteorological conditions significantly impact realized capacity, as discussed by [*22*], which is based on Bayesian Networks and can achieve prediction accuracy of 84% for departure delay times.

Some approaches to airport capacity management are not focused on determining an explicit capacity quantity, which will be used by human decision-makers, but more on developing automatic control rules, which maximize whatever capacity exists. Usually, these models are focused on a single airport and use local delay indicators to regulate traffic and mitigate the effects of delay, such as [*23*,*24*]. In these models, delay measurements at key points are used to manage the time that the aircraft spends at parking spot, before pushback, to avoid congestion in taxiways or at the runway queue.

A comprehensive approach to airport capacity, which includes representation, estimation, and optimization, is discussed in [30]. This approach emphasizes the importance of realistic capacity estimates and the dynamic allocation of airport capacities between arrivals and departures to best satisfy traffic demand. The method considers arrivals and departures as interdependent processes and uses historical data to derive practical capacity curves, which can then be used to optimize airport operations and reduce congestion.

### *B. Airspace Capacity*

Airspace capacity is a more abstract concept than airport runway capacity, because traffic in a 3-D airspace sector has more degrees of freedom than in an airport runway system. The key metrics for measuring and improving airspace capacity is airspace complexity [*25*], which has been explored intensively in literature and industry practice. An analytical model based on complexity supports the FAA Monitor Alert Parameter (MAP) [*26*,*27*]. The fundamental ideas behind it are that the airspace and traffic complexity has high influence in controller's workload, and controller's workload is the single most important factor in airspace capacity.

NASA proposed a metric that includes both traffic density (a count of aircraft in a volume of airspace) and traffic complexity (an assessment of aircraft conflicts and the difficult to solve them), referred to as Dynamic Density (DD) [*28*]. Several variants of DD were developed to accurately predict sector density and complexity [*29*,*30*,*31*,*32*,*33*]. With DD, the capacity of an airspace sector is defined in terms of the maximum complexity acceptable. Timely and accurate prediction of imbalances between capacity and demand can help traffic managers to make assertive decisions and optimize airspace resources.

An independent approach to capacity modelling is proposed in [*34*], which relates network flow theory to controller workload, providing insights on how the route structure can help to improve capacity. Another approach is based on kinetic wave theory [*35*], which observes the relationship between safety and throughput and helps evaluating conflict detection methods. More recent approaches have been exploring Machine Learning (ML) methods to approach capacity, either directly or indirectly [*36*]. Among these, [*37*] has a strong focus on capacity and proposes a completely data-driven approach for identifying the factors most influencing capacity.

### *C. Drivers for developing new capacity models*

The aforementioned models provide valuable insights into airport and airspace capacity. However, they have some shortcomings:

**Data limitations**: Most of the existing models do not explore the full potential of historical data, which impairs their ability to adapt to changing trends and patterns in air traffic operations. Some models use historical data for Data Envelopment Analysis (DEA), such as [*38*], but ignore the temporal patterns of demand and environmental conditions. Exceptions to this property are the models [*22*,*23*], which utilize real operations data, but have other limitations commented below. The lack of historical data usage results in less accurate and less reliable predictions.

**Model Complexity**: High computational requirements and/or the need for large efforts in input data collection and pre-processing can limit the scalability and applicability of these models. [*14*] requires extensive knowledge on the airport layout and utilization rules, and [*15*] requires effort to translate airport features into queuing models. [*22*] does not address automated pre-processing of input data, and the resulting model is complex and is not easy to update. The models based on DD [*29*,*30*,*31*,*32*,*33*,*28*] require the collection of many input variables which are not usually available in the operational environment, implying the need for considerable investment; thus, approximate measures have been studied [*39*,*40*] as alternatives to DD.

**Integration Challenges**: Existing models are not designed to integrate diverse data sources effectively, leading to incomplete or inaccurate predictions. Conversely, they may not produce data that is useful for the larger traffic management system, as is the case of [*23*], which addresses airport flow control needs locally, but does not address capacity measurement directly, which is a key information for the larger traffic management system.

**Model Robustness**: a model such as [*13*] heavily depends on knowing and controlling the flight schedules. In practice, schedules have considerable uncertainty and, although that model accounts for schedule uncertainty as per queuing theory, the non-stationary nature of Air Traffic makes this type of modelling highly time-consuming for researchers [*41*,*42*].

Whatever the solution developed for new capacity model, it can greatly benefit from the huge quantity of traffic data which is currently available from several sources.

## III. Background Research on Big Data Management

The System Wide Information Management (SWIM) architecture is pivotal in modern Air Traffic Management (ATM), providing standardized data services that support both operational and research activities [*43*,*44*,*45*,*46*,*47*]. SWIM data services from the FAA offer flight information services, including flight plans, aircraft positions, trajectories, and flow control messages. However, we often find that the data is uncorrelated, noisy, and voluminous, necessitating an efficient data management solution. Furthermore, there are very few providers that store and offer historical SWIM data.

We investigated various existing systems aimed at handling large-scale georeferenced data, including traditional database engines like BerlinMOD [*48*], PIST [*49*], and TrajStore [*50*], as well as distributed spatial analytics systems like Simba [*51*] and SpatialHadoop [*52*]. However, these systems either have inefficient query languages or are not specifically designed for aviation data. Systems like CloST [*53*], Elite [*54*], PARADASE [*55*], and UlTraMan [*56*] offer more efficient query processing but still lack integration for trajectory data analytics. Systems based on Apache Spark [*57*] support efficient streaming, transactions, and interactive analytics, but do not provide flexible operations and optimizations for trajectory applications.

To address these gaps, we developed a novel data management and analytics system using SWIM big data, as published in [*58*]. This system includes a data processing framework that ingests raw SWIM data into a database called R-SWIM, which is then pre-processed into a structured and indexed database called P-SWIM using MongoDB servers. The pre-processing involves a meta-preparation process with five steps: indexing, initial collection creation, data population and parsing, internal correlation, and feature extraction. This structured approach ensures efficient query performance and adaptability to new data analysis use cases. In the present paper, we generalize that methodology and present two use cases, one for predicting airspace capacity, and one for airport capacity.

## IV. Aircraft Count Prediction Capability

One of the most fundamental capabilities of our system is to predict the number of aircraft utilizing a certain airspace resource over a definite period of time. Input data to train this capability can be found in FAA's TFMS (Traffic Flow Management System), a.k.a. TFMData [*59*], which is a SWIM-compliant service that gathers and standardizes flight event data. The events of our interest are those signaling a flight entering and exiting an *airspace sector*, the definition of which being deceivingly simple. An airspace sector is a 3-dimensional volume of airspace, but when this volume is located and circumscribed in a way to contain a single airport, it will actually represent the utilization of that airport, which is a resource very distinct from airborne sectors.

Fig. 1 shows airspace sectors in the contiguous United States (US) territory, also known as CONUS. It contains a total of 1,534 airspace sectors, several of them vertically stacked, thus the figure is only notional. The color scale represents the predicted aircraft count for a sample period of time, with warmer colors (e.g. red, orange) being the sectors with higher aircraft counts, while the colder colors (e.g. green tones) being sectors with low aircraft counts.

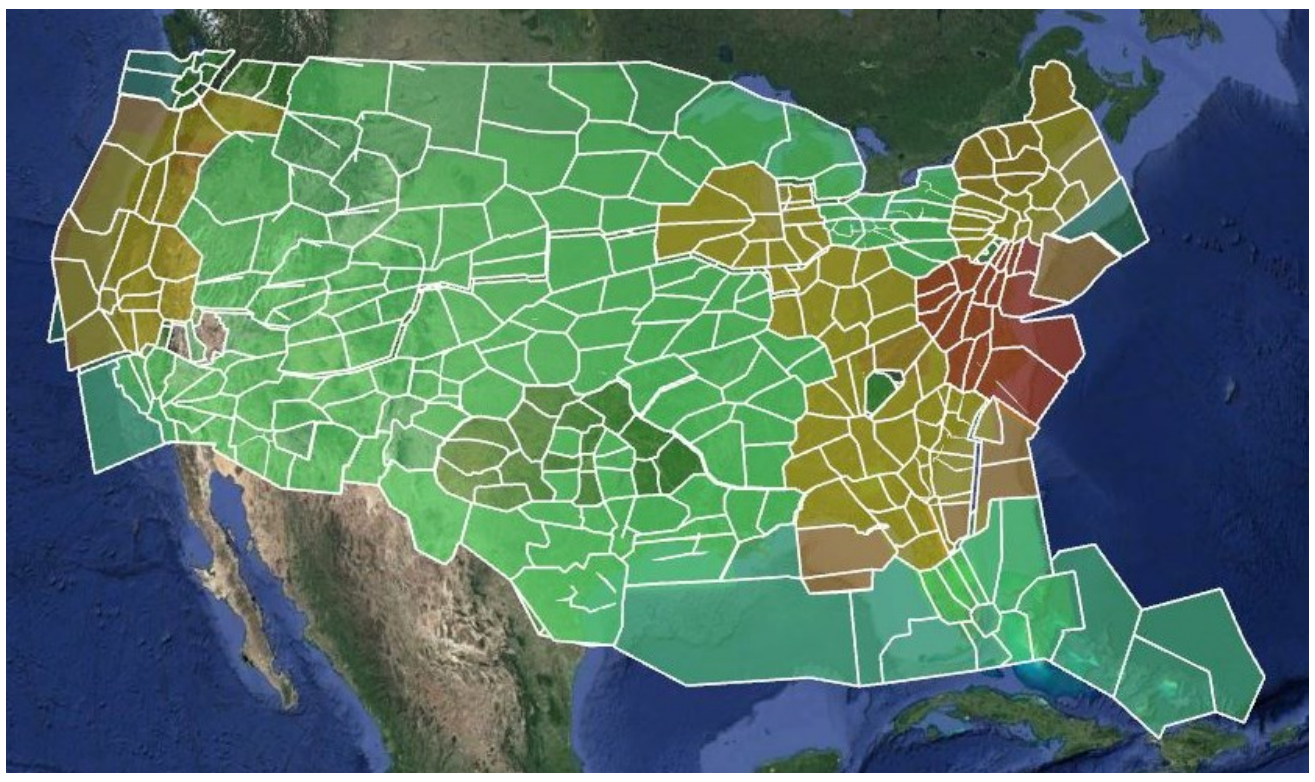

Fig. 1: Horizontal view of CONUS airspace sectors, with aircraft count represented by the color scale.

### A. Data preparation for aircraft count

A data processing framework [*58*] was developed to generate aircraft counts per sector over regular time buckets of 15 minutes each. The adapted pipeline is executed each 24-hour data set collected into R-SWIM data, initially creating a set of tables in P-SWIM, then updating it for each subsequent daily collection. This set of tables is illustrated in Fig. 2.

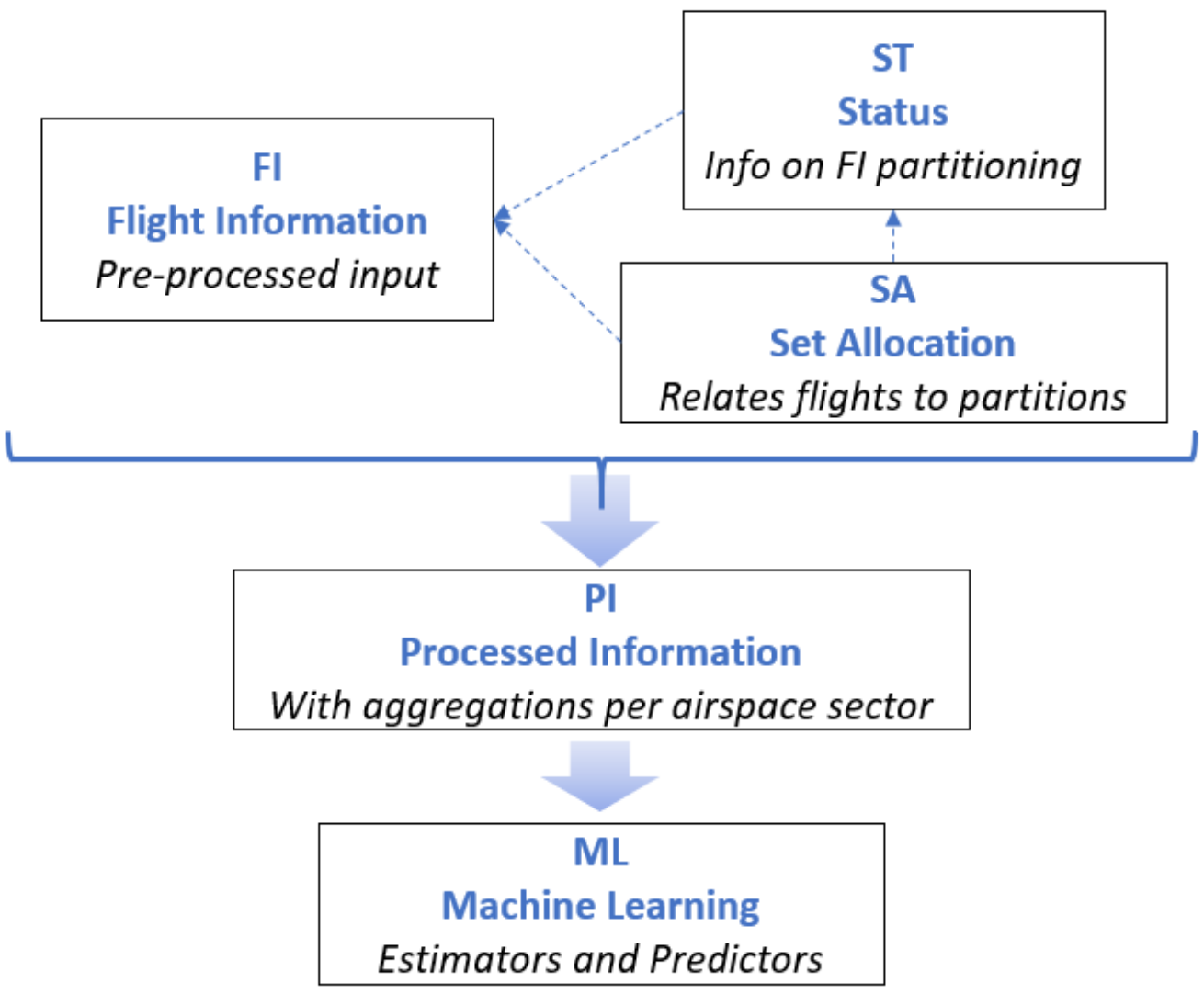

Fig. 2: P-SWIM overview of collections.

In this data schema, FI is a set of collections containing the basic flight track information, properly filtered and reconciliated, while ST helps to keep manageable size partitions of FI, according to certain rules; and SA contains indices that make queries to FI more efficient. PI contains aggregated information about flights and is organized in daily collections, and ML contains data on trained models for estimation of future values of aircraft sector count.

### B. Aircraft sector count regression model

Our regression problem consists of predicting the number $h_s$ of aircraft occupying a certain airspace sector $s$ during a time interval of a pre-defined duration. Our input data is defined as $\bar{x}_{s,p}^T = [\bar{x}_t \quad \bar{x}_w]_{s,p}$, where:

- $\bar{x}_t$ is the vector of time features: date and time (UTC);
- $\bar{x}_w$ are atmospheric weather data features: temperature, wind speed, wind direction, humidity, and pressure, occurring at time *t*;
- $p$ is the index of an individual sample ($p = 1, \dots, P$) belonging to the *P*-sized training set.

The regression model of sector $s$ must compute $h_s(\bar{x}_p)$. As the weather is univocally determined by $t$, $\bar{x}_p$ can be expressed as $\bar{x}_p(t)$ for higher clarity.

During the validation process, we found a disparity between a small number of highly populated airspace sectors, and a large number of scantly populated sectors, which was smearing out accuracy estimation. Thus, we devised a custom accuracy metric with a balancing mechanism, defined as:

$$A_s = \frac{1}{N} \sum_{t=1}^{N} e^{-\frac{|y_t - h_s(\bar{x}_p(t))|}{\text{mean}(y)}} \qquad \text{(Eq. 1)}$$

In this formula, $y_t$ are ground-truth aircraft sector count observations composing the array $y$ of sector $s$, where $t$ univocally determines the time bucket in the input data. In order to estimate the prediction accuracy with the actual data available, we use the *k*-fold cross-validation approach, which subdivides the dataset into *k* subgroups of randomly chosen samples. One of these subgroups is used as a validation set, while the remaining ones are used as training set, in order to mitigate bias that a particular split can cause. This process is repeated *k* times, each time using a different partition between training and validation data. In the case of $k = 5$, which was our choice, the evaluation uses 20% of the input dataset as test data in each fold. At the end of the process, the mean of the computed error metric is averaged to produce a performance index for the estimator.

Because we used the Scikit-learn library [*60*], the following ML algorithms were tested: Multilayer Perceptron, KNN, Bagging Classifier, Gradient Boosting, Extra-trees and Random Forest. We ran these models in parallel and evaluated their performance with the scoring measure of Eq. 1 above, with 172 days worth of data, between February and August 2018. Among these experiment runs, the Gradient Boosting Machine (GBM) [*61*] stood out with the best performance.

### C. Results

We applied our regression model to the CONUS airspace sectors, as shown in Fig. 1, with. 1,534 airspace sectors, and the 172-day data sample in P-SWIM. As each sector has its own predictor, the accuracy obtained with our ML models varied significantly according to the actual sector occupancy, i.e., the quantity being predicted. This can be visualized by means of Fig. 3.

In that figure, the horizontal axis shows the average actual daily aircraft count of each sector, and the vertical axis shows the resulting accuracy. The black plots are the direct evaluation of Eq. 1, while the green plots show that value balanced with a so-called *uncertainty feature,* which takes into account the fact that, for some of the messages in the input data, the order of the actual flight events was ambiguous. This feature makes that, in the ambiguous cases, several (up to three) ground truth values are used for the same observation, and the corresponding scores are averaged out.

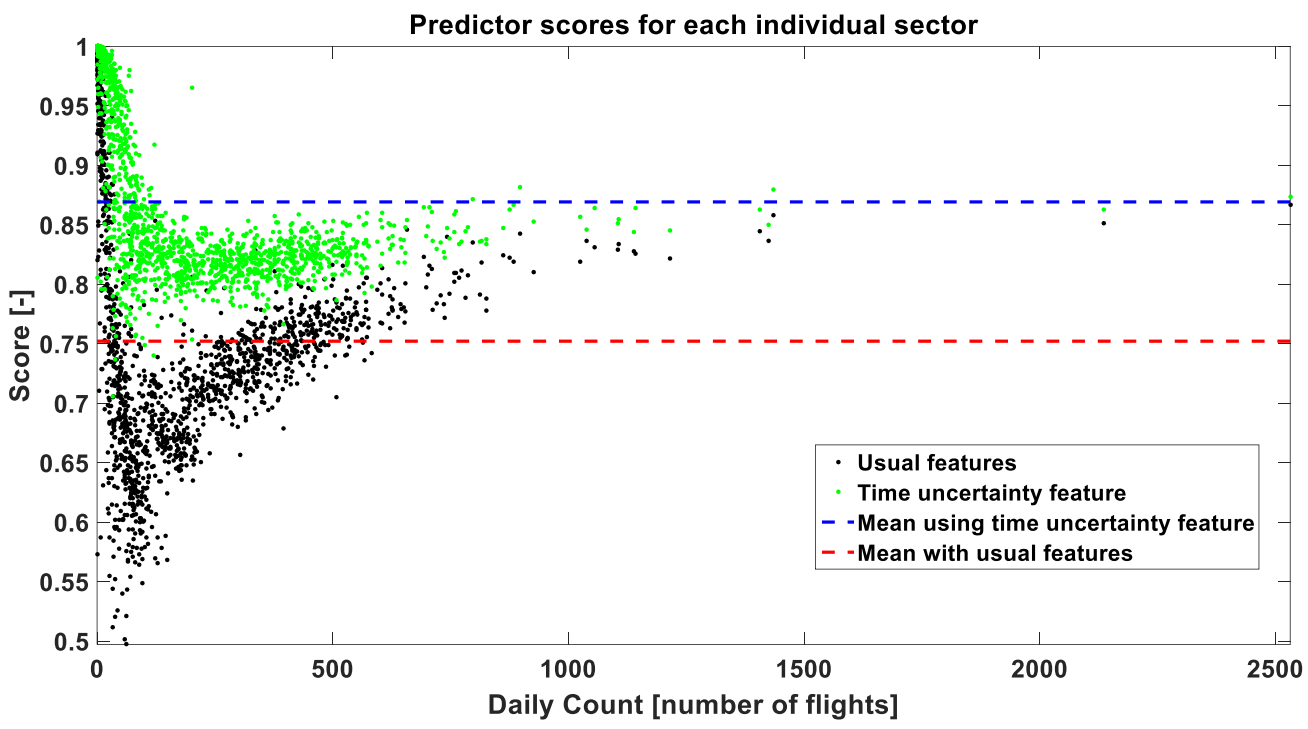

Fig. 3: Plots of score per sector, with sectors horizontally positioned by their average daily count.

It is possible to observe a trend curve among the dots, in the form of a checkmark. The least busy sectors, on the left-hand side, have a high accuracy, while the intermediary busy sectors, roughly between 100 and 200 daily flights, have the lowest accuracy, and the busiest sectors, above 450 flights, have high accuracy. The most probable explanation for this phenomenon, in our assessment, is the following: for the low-counting sectors, the ML models achieve high scores by just guessing that no flights will occur; while for the high-counting sectors, there is a large number of positive observations, which actually provide a good training for the models. On the other hand, none of this happens for the mid-counting sectors, hence their low accuracy.

## V. Airport Capacity Prediction Capability

### A. Airport Capacity Model

The second case study using our SWIM Big Data infrastructure is for airport capacity prediction. It builds upon the Sector Traffic Density Prediction Service to obtain predictions of incoming and outgoing aircraft to and from an airport. This principle is illustrated in Fig. 4.

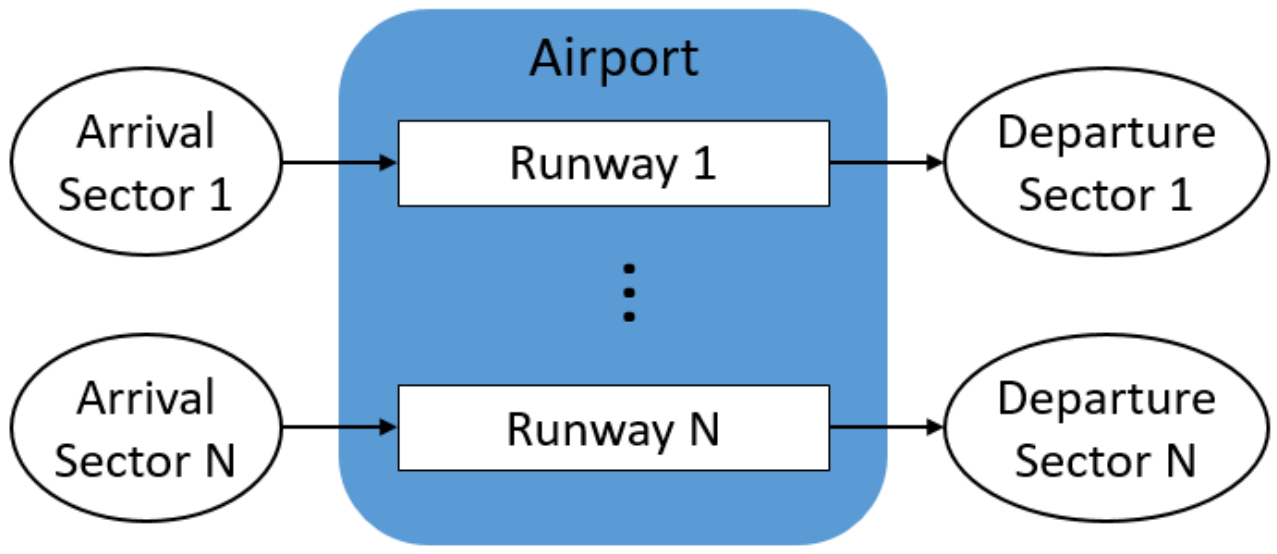

Fig. 4: Sector-Airport connectivity.

This is a simplified illustration, to convey the essential message that an airport may have one or more runways, each one of them connected to an arrival sector and a departure sector. Depending on the specific airport and airspace configuration, some arrival sectors may be common to more than one runway, as well as for the departure sectors. Also, for small airports, the arrival and departure sector may be the same. By this principle, predicting airport arrival capacity is equivalent to predicting the sum of the output rates of the respective arrival sectors and, conversely, predicting airport departure capacity is equivalent to predicting the sum of intake rates of the departure sectors.

There are official regulations constraining airport capacity in low visibility and severe weather conditions [*16*,*17*,*18*,*19*], so weather variables are part of the inputs. Besides weather radar data, there are periodical reports of current and forecast weather that are based on measurements taken by weather balloons in the airspace directly above or in the surroundings of the airport's perimeter. Processing this data results in the most widely used weather reports, called METAR (METeorological Aerodrome Report) and TAF (Terminal Area Forecast) [*62*]. These reports are issued at pre-defined regular times, with METAR containing the present weather, and TAF containing the forecasted weather changes until a given time horizon. Thus, the high-level data flow of our Airport Capacity Prediction Service (P.S.) becomes as in Fig. 5.

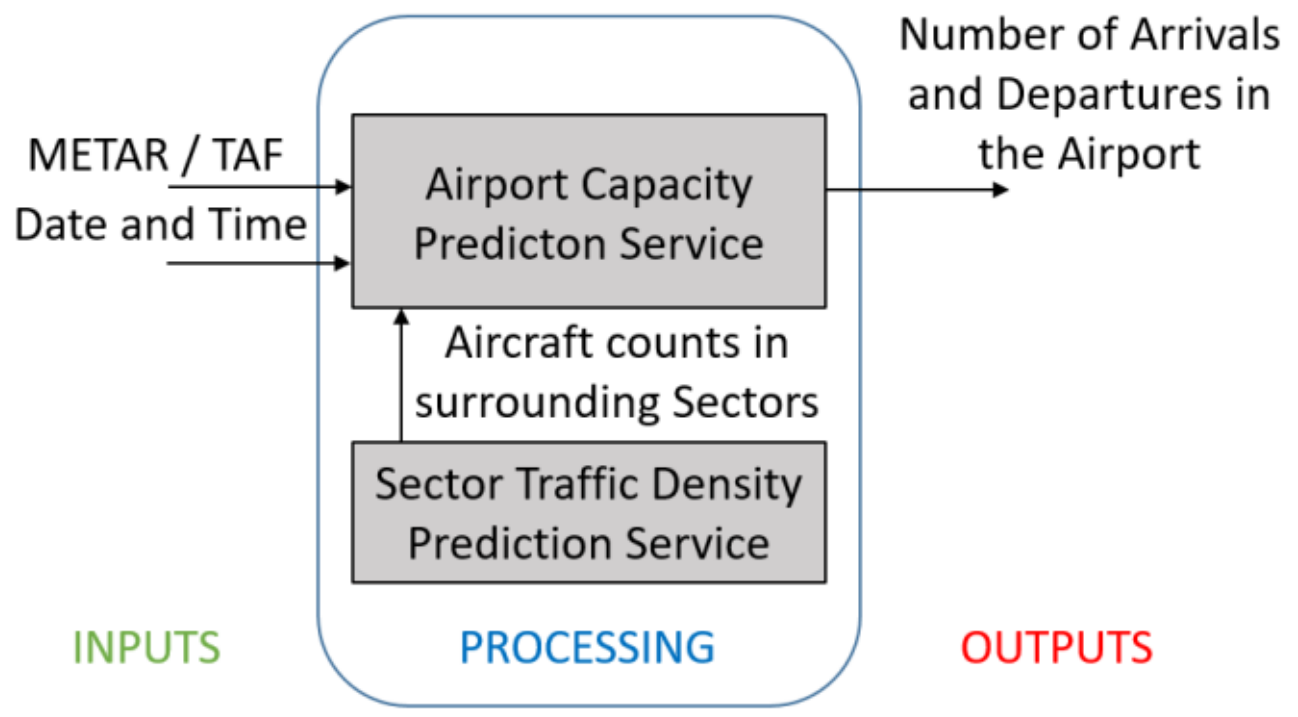

Fig. 5: High-level dataflow of the Airport Capacity Prediction Service.

According to this dataflow, Airport Capacity P.S consumes the predicted aircraft counts of the surrounding sectors from the Sector Traffic Density P.S. (described in Section IV), which thereby must pre-exist. Airport Capacity P.S. also consumes METAR and TAF data, and issues predictions of the number of aircraft arrivals and departures for a series of time windows in the future. For airports, where historical data on aircraft movements per runway is available, it is possible to issue Arrival and Departure counts per runway, but in our model's outputs, we only show airport-level predictions, and runway data is used only internally.

Before a model of Airport Capacity is built and trained, our approach builds and trains a predictor of airport Runway Configuration (RC), given that this factor is a major determinant of airport capacity. Despite an airport may have many runways, their activations and senses are configured according to a small number of combinations, known by the airport tower managers. These configurations, in turn, are strongly determined by the wind direction and visibility conditions [*63*]. Thus, the causal order of the phenomena is: weather → RC → airport capacity. However, from a broader perspective of National or Multi-National traffic management, the RC of an airport is not as relevant as the overall airport capacity.

Also, we used only METAR data in our prototype implementation, because all our tests were done with past data, thus the use of TAF would not add significant value. The advantage in processing TAF data would be significant in a model that incorporates the present traffic and weather situation to issue predictions within a varying time horizon, which is not the case of our prototype. Similarly, to the Sector Count P.S., our Airport Capacity P.S. is stationary with regard to scheduled traffic data, which is not used as input.

### B. *Airport Capacity Prediction results*

To build and train this predictive service, we augmented the P-SWIM database, explained in the previous section, with European SWIM data, focused on the airspace sectors surrounding the Frankfurt (FRA) airport, with respective METAR data. The results from training, selecting and testing the ML predictor, are shown on Fig. 6 and Fig. 7.

The validation, testing and evaluation of the Airport Capacity prediction model was done by the same method employed for the Sector Traffic Density prediction model (Subsection IV). Similarly, all Scikit-learn model types were evaluated, and the top performing two models were Gradient Boosting (GB) and Logistic Regression. Overall, GB was selected as the best performing model.

Fig. 6 shows the temporal series of actual and predicted values in a certain day of the sample, where it can be seen that the predictions are a plausible approximation of the actual values. Fig. 7 shows the histogram of scores, calculated according to Eq. 1, for all time buckets of 15 minutes in the sample, which comprised 26 days. Both arrival and departure predictions exhibit mean accuracy scores around 80%.

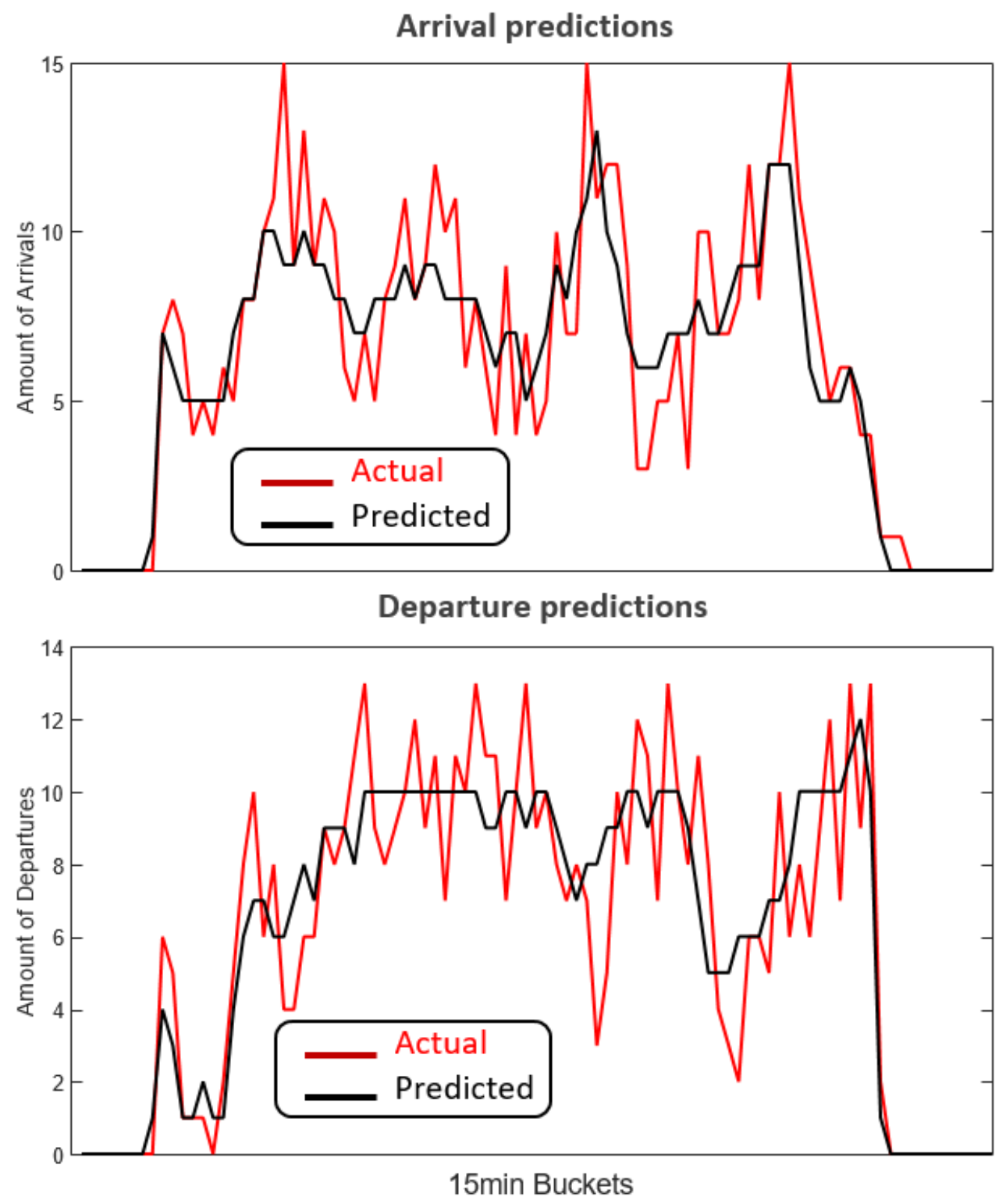

Fig. 6: Arrival and Departure prediction time series for the FRA airport.

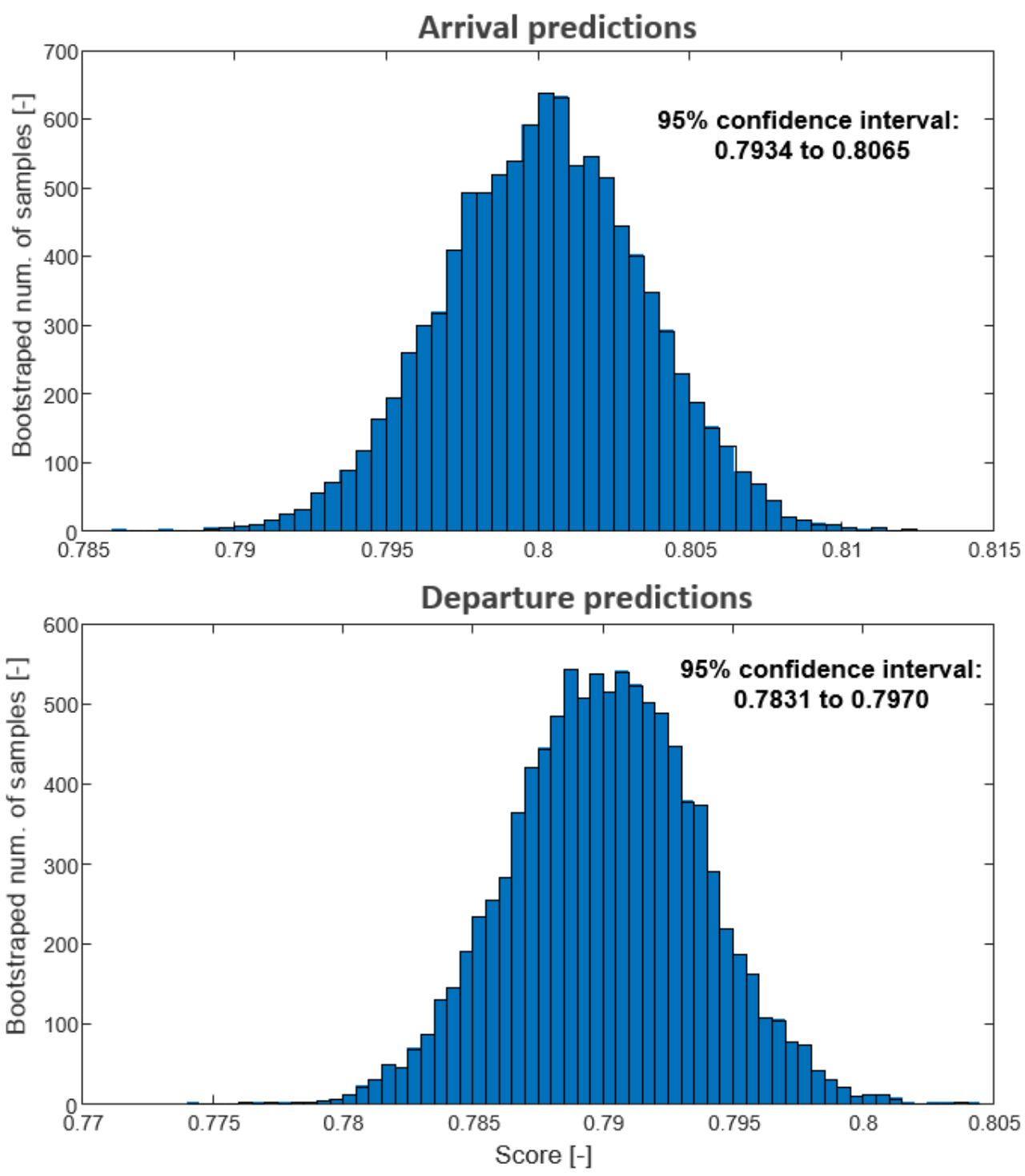

Fig. 7: Distribution of prediction scores of the Airport Capacity model.

## VI. Compositional Services Architecture

The Airport Capacity case study demonstrated that prediction services can use input from more elementary prediction services, also referred to as micro-services, in a compositional fashion. Actually, National and Multi-National predictions of congestion states may be issued by Meta-Services at a higher layer, consuming input from lower-level services. An example of such type of architecture is depicted in Fig. 8.

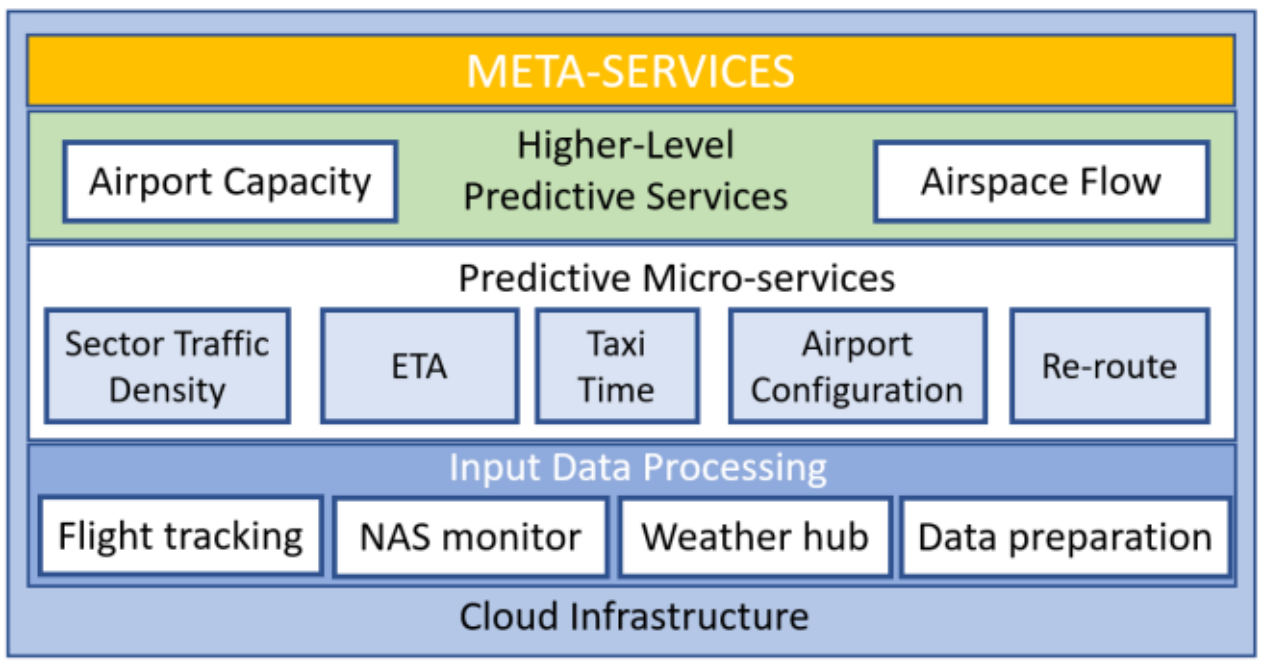

Fig. 8: Compositional Services Architecture based on Micro-Services.

At the lowest and most pervasive layer of the architecture, the Cloud Infrastructure provides database management systems, message queues, networking interfaces, security and other common IT resources. The Input Data Processing layer collects data from external data sources, and executes data preparation pipelines, as explained in Section III. With the pre-processed data, the Predictive Micro-Services of the next layer can already train basic Machine Learning models. The distinction between Micro-Service and Higher-Level service is not crystalized, but in general we may say that a Micro-Service does not have a direct interface with the end-user. A Higher-Level service may either compute its predictions by direct execution of stationary algorithms or AI/ML models of their own, fed by data from the lower-level services. Also, some of the lower-level services may be unavailable and the respective consuming services may be tolerant to these faults. More definitions about this architecture can be found in [*64*].

We demonstrated the feasibility of such hierarchical architecture by having Sector Traffic Density P.S. and the Airport Configuration P.S. providing input to the Airport Capacity P.S., and developing the Re-route prediction service, as presented in [*65*]. While we developed an interface that was capable of retrieving data from a commercially available Taxi Time P.S. [*66*], we did not have resources to enhance our model to adequately consume this input. This was similar to the ETA (Estimated Time of Arrival) P.S., with the difference that the ETA predictor [*67*] was not a commercially available tool.

## VII. Other Categories of Predictive Capabilities

This paper is not an exhaustive survey, thus there may still be a large number of references containing diverse types of predictive capabilities in air transportation. In particular, delay prediction is a topic that has rendered the development of remarkably effective solutions. Despite a direct comparison with our predictive services is not possible, those interested in knowing state of the art delay prediction algorithms can find good references in [*68*,*69*,*70*,*71*].

## VIII. Final Remarks

Our novel SWIM data processing infrastructure is capable of ingesting large volumes of aeronautical data, and training Machine Learning models capable of predicting airspace sector congestion, airport capacity and reroutes. They share the principle of observing temporal series of input variables, in which each observation is a summation or aggregation of occurrences of certain events within a fixed time bucket, paired to meteorological data and, in theory, any other data which may be correlated with the predicted variables.

The strength of our approach is to be virtually unlimited in the prediction horizon. Given the airspace system resource in question, a time bucket in the future, defined by date and time, and a weather forecast for that time bucket, our models are capable of predicting measures and events of interest with accuracy scores above 80%, without looking at flight schedules. We believe that predictive capabilities like this can greatly improve the efficiency of ATM around the globe, by helping initiatives involving proactive management of capacity balance, such as dynamic re-sectorization [*72*], Collaborative Trajectory Option Set (CTOP) [*73*,*74*], and several others.

There are several directions of work that would help to increase the benefits offered by this system, and its overall applicability in practical settings:

- **Processing real-time traffic data**: the states of congestion are strongly correlated in the span of a few hours, as well as the weather conditions. Currently, our model is capable of daily updates, however the models give equal importance to yesterday and any other day in the past, ignoring temporal proximity. In order to properly account for different temporal importance, our models would have to use Recurrent Neural Networks (RNN) or Long Short-Term Memory (LSTM) networks.
- **Processing flight schedules**: airline flight schedules are a strong determinant of traffic volumes in the mid and short term, between a few months and the flight departure time.

Thus, they also can improve the model accuracy in these time horizons.

- **Identification of which flight plans will be directly affected by a reroute and how they will be affected**: this is possible when having information on scheduled and confirmed flight plans for the near future. But, beyond simple association, there may be indirect impacts due to congestion, and furthermore it would be desirable to automatically propose new flight plans that are compatible with the reroute advisories. A data-driven approach for these capabilities sounds the most promising way forward.
- **Provision of dynamic confidence intervals for prediction scores**: it would be useful to output confidence intervals for the prediction scores, so that, when making a decision, the user could tune her/his reliance on the prediction model.
- **Integration of Reroute Prediction to the CTOP process**: by having reroute prediction, it is possible to compute revised flight plans that are compliant with the reroute advisories and, as a logical consequence, these solutions could be leveraged by integration with CTOP.
- **Using High-Performance Computing (HPC) to increase predictive power**: as shown in the paper, the present system does a lot of aggregations in the input variables in order to keep down the number of inputs and be manageable in a regular notebook computer. If we assume an HPC platform with powerful GPU units, large memory, and dozens of CPUs, this enhanced computing power allows more sources and more years of data to be used in the training, with less aggregation, and that would supposedly lead to better performance in the difficult cases. The anticipated proposal of alternative routes would most certainly require HPC resources.
- **Extension to other regions of the world**: wherever aviation traffic data can be collected, the same principles of this system can be applied. For example, the Eurocontrol Network Manager has the B2B rerouting service and is a clear possibility. But there may be other cases that we have not started to explore.
- **Exploration of more efficient databases**: although MongoDB has been evolving and providing enough performance as part of our infrastructure, other database engines might contribute to further improvement in our system's performance. Just to cite an example, the OpenSky flight tracking data service utilizes the Trino database engine for historical data [*75*].